\documentclass[letterpaper, 10 pt, conference]{ieeeconf}  

\IEEEoverridecommandlockouts                              

\usepackage{graphics} 
    \DeclareGraphicsExtensions{.jpg,.pdf,.mps,.png} 
    \graphicspath{{fig/} {./}} 
\usepackage{amsmath} 
\usepackage{amssymb}  
\usepackage{setspace} 
\usepackage{bm} 
\usepackage{flushend} 

\newcommand{\fig}[1]{Fig.~\ref{#1}}

\newcommand{\eq}[1]{Eq.~(\ref{#1})}

\def\epsgaiji#1{\leavevmode\kern-0.025zw\raise-.37zh\hbox{%
  \epsfile{file=#1,width=1.05zw}}\kern-0.025zw}
\newcommand{\MARU}[1]{{\ooalign{\hfil#1\/\hfil\crcr\raise.167ex\hbox{\mathhexbox20D}}}}

\usepackage{array}

\usepackage{siunitx} 
\usepackage{gensymb} 
\usepackage{multirow} 
\usepackage{caption}
\usepackage{subcaption}
\usepackage{colortbl} 
\usepackage{xcolor} 

\usepackage{soul}
\usepackage{framed} 
\usepackage{svg}

\usepackage{pgfplots}
\pgfplotsset{compat=newest}
\usetikzlibrary{plotmarks}
\usetikzlibrary{arrows.meta}
\usepgfplotslibrary{patchplots}
\usepackage{xcolor}

\definecolor{mygreen}{rgb}{0.0,1.0,0}
\definecolor{mylightgreen}{rgb}{0.7,0.9,0.0}
\definecolor{myyellow}{rgb}{1.0,1.0,0.2}
\definecolor{myorange}{rgb}{1.0,0.5,0}
\definecolor{myred}{rgb}{1.0,0,0}

\usepackage{grffile}
\pgfplotsset{plot coordinates/math parser=false}
\newlength\fwidth
\newlength\fheight

\usepackage{cite}

\title{\LARGE \bf
Optimal Trajectory Planning for Orbital Robot \\ Rendezvous and Docking}

\author{Kenta Iizuka$^{1}$, Akiyoshi Uchida$^{1}$, Kentaro Uno$^{1}$ and Kazuya Yoshida$^{1}$
\thanks{$^{1}$Kenta Iizuka, Akiyoshi Uchida, Kentaro Uno, and Kazuya Yoshida are with the Space Robotics Lab. (SRL) in the Department of Aerospace Engineering, Graduate School of Engineering, Tohoku University, Sendai 980-8579, Japan.}%
\thanks{\textit{Corresponding author is Kenta Iizuka.}}%
\thanks{~~~~(Email: \texttt{\small iizuka.kenta.r5@dc.tohoku.ac.jp})}%
}

\begin{document}

\maketitle
\thispagestyle{empty}
\pagestyle{empty}


\begin{abstract}

Approaching a tumbling target safely is a critical challenge in space debris removal missions utilizing robotic manipulators onboard servicing satellites. 
In this work, we propose a trajectory planning method based on nonlinear optimization for a close-range rendezvous to bring a free-floating, rotating debris object in a two-dimensional plane into the manipulator's workspace, as a preliminary step for its capture.
The proposed method introduces a dynamic keep-out sphere that adapts depending on the approach conditions, allowing for closer and safer access to the target. 
Furthermore, a control strategy is developed to reproduce the optimized trajectory using discrete ON/OFF thrusters, considering practical implementation constraints.

\end{abstract}



\section{Introduction}
\label{sec:intro}
\subsection{Background}
\label{subsec:background}

In recent years, the amount of space debris in Earth's orbit—such as defunct satellites and upper stages of launch vehicles—has been rapidly increasing, posing a growing threat to active satellites and crewed spacecraft\cite{esa_spacedebris}. 
These debris objects travel at extremely high speeds of several kilometers per second, and any potential collision can result not only in the destruction of operational spacecraft or mission failure, but also in the generation of additional debris, potentially triggering a cascade effect known as the Kessler syndrome\cite{kelso2009, kessler2010}. 
Against this backdrop, the importance of active debris removal through on-orbit servicing (OOS) has gained widespread attention, and research and development efforts are being pursued globally.

\begin{figure}[t]
    \centering
    \includegraphics[width=\linewidth]{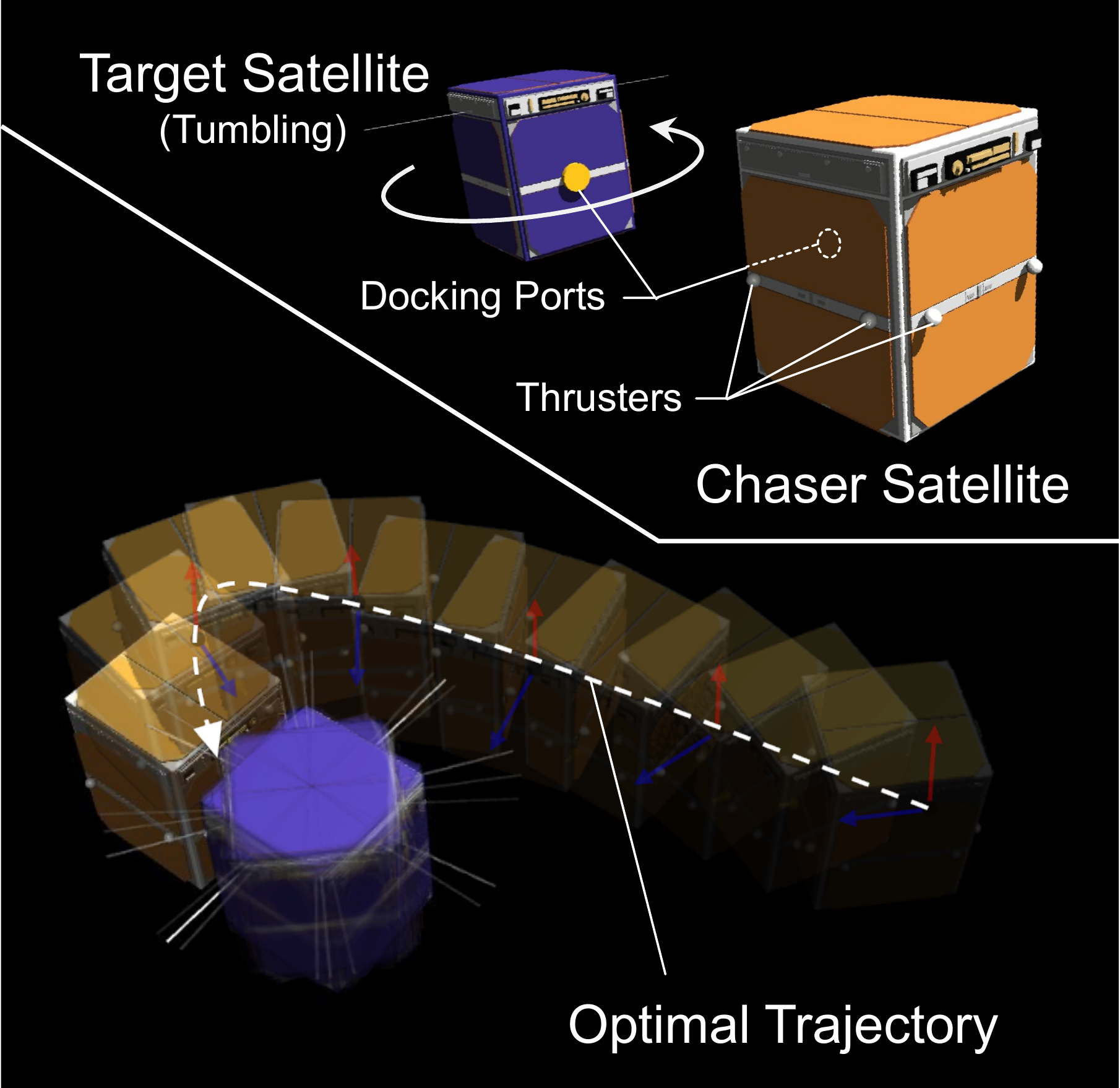}
    \caption{Simulation of a thruster-mounted chaser satellite approaching a target satellite in rotational motion (top-right) and simulated approach trajectory of the chaser achieving a precise adjustment of the pose (bottom).}
    \label{fig:MuJoCo_1}
\end{figure}

Various methods for space debris removal have been proposed and investigated~\cite{shan2016}. 
Among them, contact-based capture using robotic manipulators has attracted particular attention due to its ability to achieve precise grasping and attitude control of the target~\cite{papadopoulos2021}, in contrast to other methods such as tethers or harpoons~\cite{wormnes2013,ishige2004}. 
Moreover, manipulation-based capture is not limited to mere debris collection; it also holds promise for future high-value OOS missions such as satellite repair and propellant refueling, demonstrating its versatility as a general-purpose approach. 
In addition, experimental demonstrations of debris capture using actual hardware in two-dimensional microgravity environments have also been conducted~\cite{uchida2024}, supporting the feasibility of contact-based approaches.

Debris removal using robotic manipulators generally consists of the following steps:  
(1) observation and motion estimation of the target,  
(2) approach to a proximity range where the target enters the manipulator's workspace, and  
(3) final capture by the manipulator.  
Among these phases, the second one—approaching the target—is particularly challenging, especially when the target is tumbling. 
In this phase, the servicing satellite must safely approach a close distance while avoiding collision, making trajectory planning one of the most difficult aspects of the mission.

In recent years, increasing attention has been given to trajectory planning methods tailored specifically for the approach phase of rendezvous. 
Among them, optimization-based approaches have been widely proposed, in which the predicted motion of the target and various safety constraints are incorporated into a formal optimization framework to generate feasible approach trajectories. 
However, these methods often need to handle nonlinear dynamics, actuator limitations, and collision avoidance constraints at the same time, making it difficult to maintain consistency between theoretical formulation and practical implementation.

As an example of an optimization-based approach, Albee et~al.\ proposed a robust observation, planning, and control pipeline employing Tube-based Model Predictive Control (Tube-MPC) to address tumbling targets~\cite{albee2021, specht2023}.
While their method demonstrates high robustness, it focuses on generating continuous-valued ideal trajectories and does not address the practical challenge of ON/OFF thruster control, which is commonly used in actual spacecraft systems.

As an alternative to optimization-based methods, potential field-based approaches are also representative~\cite{zappulla2019}, among which the Improved Disturbed Fluid Method (IDFM) is one notable extension~\cite{wang2023trajectory}. 
In these studies, the target is often assumed to be stationary. 
Chen et~al.\ successfully conducted an experiment in which a chaser was maneuvered into a narrow space between the solar panels of a slowly rotating target in a two-dimensional plane, providing highly valuable insights~\cite{chen2022ground}.
Nevertheless, a fundamental limitation of potential field-based methods is that they do not directly take into account the energy efficiency of the chaser, which is of critical importance for practical applications.

Furthermore, many of these existing studies define conservatively shaped Keep-Out Spheres (KOS), such as circular or spherical regions, to ensure safety. 
While effective in avoiding collisions, such conservative KOS settings restrict the chaser’s ability to approach the target in close proximity, particularly to the level required for manipulator operations or docking maneuvers.

\subsection{Motivation}
\label{subsec:motivation}

This study aims to generate a close-range rendezvous trajectory for tumbling space debris, under the assumption that it will be captured by a robotic manipulator. 
Specifically, the objective is to bring the chaser satellite into proximity so that the debris lies within the operational workspace of the manipulator. 
The motion of the target is assumed to be known. 
This assumption is supported by prior studies showing that motion estimation methods based on CNN~\cite{price2023} and Structure from Motion~\cite{uno2024} can estimate the target's motion information with an error of only a few percent.

In this context, the feasibility and safety of the approach depend on the relative state between the chaser and the target. 
To address this, the proposed method introduces a dynamic Keep-Out Sphere (KOS) whose shape adapts based on the relative configuration, rather than relying on a static exclusion zone. 
This enables more flexible and safer trajectory planning while respecting safety constraints.

Moreover, the generated trajectory is not merely treated as an ideal reference path; rather, it is designed with practical implementation using discrete ON/OFF thrusters in mind.
To this end, the thruster usage cost is included in the objective function of the optimization, and a feedback control law is constructed to track the optimized trajectory. 
A thrust allocation method is also developed to convert the continuous wrench into discrete thruster commands suitable for PWM-based actuation. 
The overall system is validated through physics-based simulation. 
In this simulation, model inaccuracies and actuator constraints are taken into account to evaluate the practical feasibility and robustness of the approach for real-world implementation.

By explicitly modeling the system dynamics, constraints, and thruster structure in analytical form, the proposed method enables a clearer interpretation of the generated trajectory concerning its physical and structural properties. 
This, in turn, facilitates both trajectory generation and execution, maintaining consistency between theoretical formulation and practical implementation.

\subsection{Contribution}
\label{subsec:contributions}
The primary contribution of this work lies in the development and evaluation of a comprehensive framework for generating and reproducing a safe and feasible approach trajectory toward a space debris with known rotational motion.

To validate the proposed method, two types of simulations were conducted. 
First, the theoretical validity of the trajectories was verified in a simulation under ideal conditions, using the same mathematical model as the optimization. 
Second, their practical feasibility and reproducibility were assessed in a high-fidelity MuJoCo\cite{MuJoCo} simulation under realistic conditions, which incorporated imperfections such as model discrepancies and actuator discretization.

In addition, we quantitatively evaluated the trajectory generation performance under varying approach conditions, including the target’s attitude, angular velocity, and the chaser’s maximum thruster output. 
These results provide insights into the applicability and performance limits of the proposed method and contribute to the design of more efficient debris capture strategies in future on-orbit servicing missions.

\section{Preliminaries}
\subsection{System States and Dynamics}
In this study, we model the motion of a chaser and a target spacecraft in a two-dimensional plane, representing the space environment. 
The chaser is modeled as a square rigid body with translation and rotational degrees of freedom.

In the center-of-mass frame, the chaser's state is defined as a 6-dimensional vector as follows.

\begin{equation}
    \bm{x} =
    \begin{bmatrix}
    x & y & \theta & v_x & v_y & \omega
    \end{bmatrix}^\top
    \label{eq:dynamics}
\end{equation}

Here, $(x, y)$ denotes the position of the chaser's center of mass, $\theta$ is the attitude angle (orientation concerning the inertial frame), $(v_x, v_y)$ are the linear velocities, and $\omega$ is the angular velocity. 
The motion of the chaser is governed by the Newton--Euler equations:

\begin{equation}
    \begin{aligned}
    \dot{x} &= v_x, \quad \dot{y} = v_y, \quad \dot{\theta} = \omega \\
    \dot{v}_x &= \frac{1}{m} f_x, \quad \dot{v}_y = \frac{1}{m} f_y, \quad \dot{\omega} = \frac{1}{I} \tau
    \end{aligned}
    \label{eq:newton-euler}
\end{equation}

Here, $m$ is the mass of the chaser, $I$ is its moment of inertia about the z-axis through its center of mass, $(f_x, f_y)$ are the external forces, and $\tau$ is the applied torque. 
These forces and torques are generated by multiple thrusters mounted on the surface of the chaser. 
Each thruster produces thrust perpendicular to the surface it is attached to, and is controlled in a discrete ON/OFF manner with a fixed direction. 
The total force and torque applied to the chaser, $(f_x, f_y, \tau)$, are the result of the combined output of these thrusters.

The target is assumed to rotate about its center of mass with a known angular velocity $\omega_{\text{target}}$ and attitude $\theta_{\text{target}}$, while its position remains fixed. 
This assumption is motivated by previous reports indicating that, in actual space environments, large space debris often undergo slow rotation about a single axis~\cite{benson2019}.
Accordingly, this study focuses on simulations conducted on a plane perpendicular to the axis of rotation, which serves as the basis for the trajectory optimization and dynamic simulation described later.

\subsection{Thruster Model} 
\begin{figure}[t]
    \centering
    \includegraphics[width=0.8\linewidth]{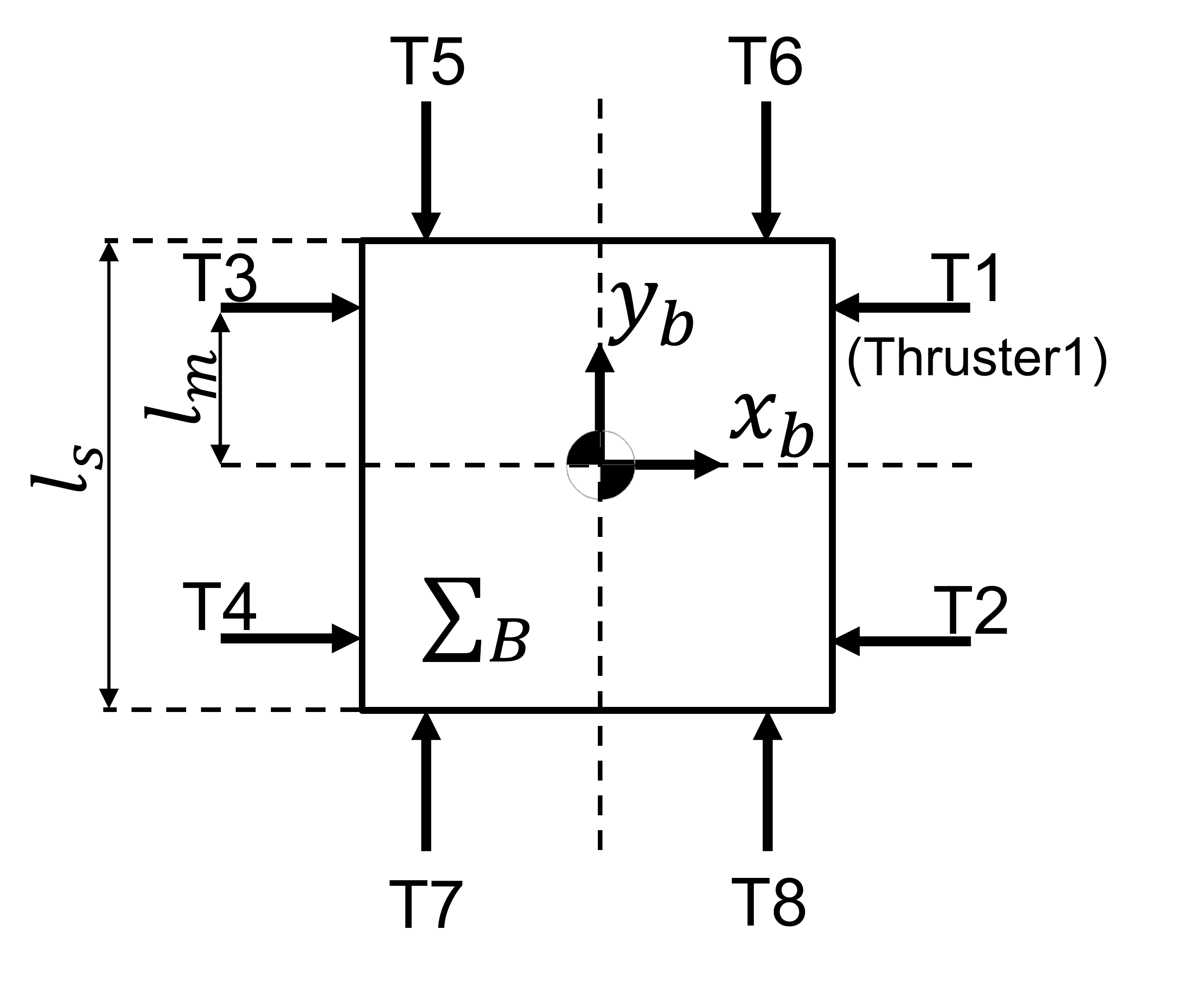}
    \caption{Placement of thrusters and direction of thrust.}
    \label{fig:thruster_model}
\end{figure}

In this study, the chaser spacecraft operates in a two-dimensional plane and controls its translational and rotational motion using eight thrusters mounted on its surface (\fig{fig:thruster_model}). 
Each thruster has a predefined mounting position and thrust direction, and its output is determined by discrete ON/OFF control. 
Specifically, whether thruster $i$ is fired at a given time is represented by a binary variable $u_i \in \{0, 1\}$.

The position vector of thruster $i$ is denoted by $\bm{r}_i = [r_{ix}, r_{iy}]^\top \in \mathbb{R}^2$, and its thrust vector is $\bm{f}_i = f_{\max} \bm{d}_i$ (the product of the maximum force magnitude $f_{\max}$ and the unit direction vector $\bm{d}_i$), where $\|\bm{d}_i\| = 1$. 
When a thruster is ON, it generates thrust $\bm{f}_i$; otherwise, it produces zero thrust. The total force and torque generated by all thrusters are given by:

\begin{align}
    \bm{f}_{\text{total}} &= \sum_{i=1}^{8} u_i \bm{f}_i \label{eq:thruster_f} \\
    \tau_{\text{total}} &= \sum_{i=1}^{8} u_i (\bm{r}_i \times \bm{f}_i) \label{eq:thruster_tau}
\end{align}

Here, $\bm{r}_i \times \bm{f}_i$ denotes the two-dimensional cross product, defined as a scalar quantity.
By substituting $\bm{f}_{\text{total}}$ and $\tau_{\text{total}}$ into the equations of motion, the chaser’s motion is fully determined by the ON/OFF combinations of the thrusters.

The thruster configuration adopted in this study is based on those commonly used in planar air-floating testbeds (e.g.,~\cite{santaguida2023, kwokchoon2018}), which enable a straightforward reproduction of both translational and rotational motion in two dimensions. 
At the same time, this setup can be regarded as a simplified planar representation of a three-dimensional spacecraft configuration—specifically, one equipped with eight thrusters mounted individually at the corners of a cubic structure.
Therefore, the proposed two-dimensional simulation ensures consistency with future air-floating experiments while maintaining structural correspondence with a practical three-dimensional spacecraft system.
Although this configuration is somewhat redundant for planar motion, the main objective of this work is to validate the effectiveness of the proposed trajectory-generation and tracking framework, rather than to optimize the propulsion system design.

\section{Method}
\subsection{Optimization Model}
In this study, the motion of the chaser is formulated as a nonlinear optimization problem to achieve safe proximity to the target, inspired by the approach in~\cite{albee2021}.
The chaser is modeled as a two-dimensional rigid body in the inertial coordinate frame.
The state vector is defined as \( \bm{x} = [x, y, \theta, v_x, v_y, \omega]^\top \), and the control input vector is defined as \( \bm{\mathcal{F}} = [F_x, F_y, \tau]^\top \). 
Here, \( x, y, \theta \) represent the position and orientation angle, \( v_x, v_y, \omega \) are the translational and angular velocities, and \( F_x, F_y, \tau \) denote the force and torque in the inertial frame.

The continuous-time problem is discretized with a fixed time step \( \Delta t \), and formulated as a nonlinear programming (NLP) problem using the forward Euler method. 
The optimization solver used was IPOPT (Interior Point OPTimizer), integrated into CasADi\cite{andersson2019casadi}.
In this study, the target's attitude at closest approach is predefined. 
Accordingly, the maneuver duration \( t \) is chosen from a set of candidate values derived from the target’s rotation period. For each candidate \( t \), the optimization is executed in ascending order of duration, and the solution that yields the best performance is selected as the final result.

\subsection{Objective Function and Constraints}
The objective function \( J \) of the optimization problem is defined in ~\eqref{eq:objective}.

\begin{align}
  \min_{\{\bm{\mathcal{F}}_k\}} J = 
  & \; w_{\text{goal}} \| \bm{x}_{N} - \bm{x}_{\text{goal}} \|^2 \nonumber \\
  & + \sum_{k=0}^{N-1} \left( E_{\mathrm{kin},k} + w_{u} \| \bm{\mathcal{F}}_k \|^2 \right) \Delta t
  \label{eq:objective}
\end{align}

\begin{align}
  & \bm{x}_0 = \bm{x}_{\text{init}}, \quad \theta_N = \theta_{\text{finish}}\label{eq:init_state} \\
    & \bm{x}_{k+1} - (\bm{x}_k + f(\bm{x}_k,  \bm{\mathcal{F}_k})\Delta t) = \bm{0} \label{eq:dynamics} \\
  &  \bm{\mathcal{F}}_{\min} \le  \bm{\mathcal{F}}_k \le  \bm{\mathcal{F}}_{\max}\label{eq:control_limits} \\
  & g(\bm{x}_{\text{chaser},k}, \bm{x}_{\text{target},k}) \ge 0 \label{eq:collision_avoid} 
\end{align}

The first term on the right-hand side of ~\eqref{eq:objective} represents the squared error between the final position of the chaser \( \bm{x}_{N} \) at the final time step \( N \) and the desired goal position \( \bm{x}_{\text{goal}} \), contributing to terminal accuracy. 
The second term is the time integral of the chaser’s kinetic energy \( E_{\mathrm{kin},k} \) at each time step \( k \), which promotes trajectory smoothness. 
The third term represents the squared control input \(  \bm{\mathcal{F}}_k \), serving as a proxy for minimizing fuel consumption.  
The weights \( w_{\text{goal}} \) and \( w_u \) are the respective weighting coefficients.

The constraints considered in this optimization are shown in ~\eqref{eq:init_state}--\eqref{eq:collision_avoid}.
\eq{eq:init_state} imposes boundary conditions that constrain the initial state and final attitude of the chaser.  
\eq{eq:dynamics} represents the discretized equations of motion, enforcing adherence to the laws of physics.  
\eq{eq:control_limits} reflects the physical limits of the forces the chaser’s actuators can apply.  
\eq{eq:collision_avoid} represents the collision avoidance constraint, where \( g \) is a distance function from the Keep-Out Sphere (KOS), defined as a forbidden region around the target’s center of mass.
In this study, the objective is to guide the chaser into the operational workspace of its manipulator, rather than to achieve strict docking solely through center-of-mass translation. Accordingly, the terminal position is treated as a soft objective within the cost function. The keep-out zone constraint is also a positional constraint, but it is retained as a hard constraint to prioritize safety.

\subsection{Keep-Out Sphere (KOS)}
\begin{figure}[t]
  \centering
  \subcaptionbox{KOS in State I.}[0.48\linewidth]{%
    \includegraphics[width=\linewidth]{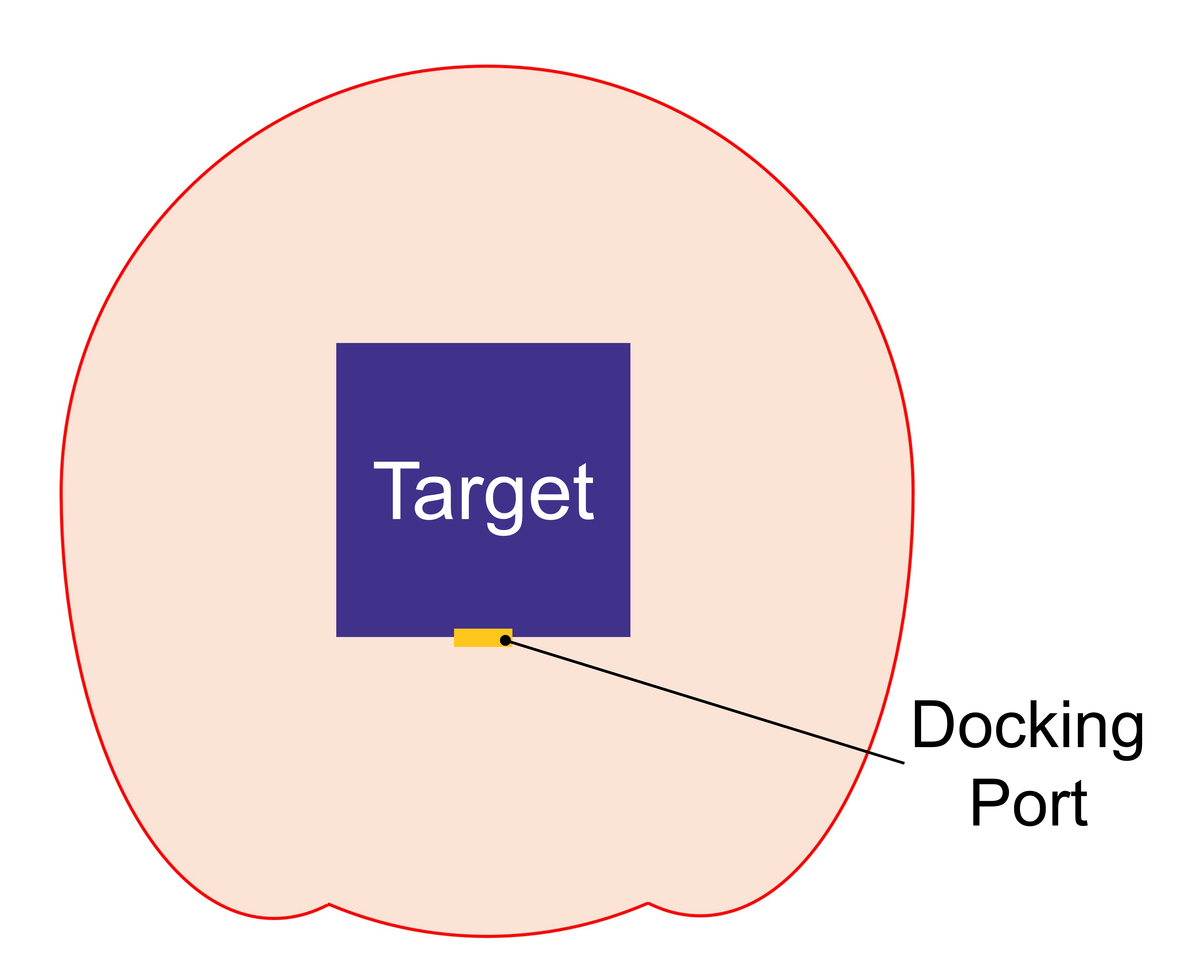}
  }
  \hfill
  \subcaptionbox{KOS in State II.}[0.48\linewidth]{%
    \includegraphics[width=\linewidth]{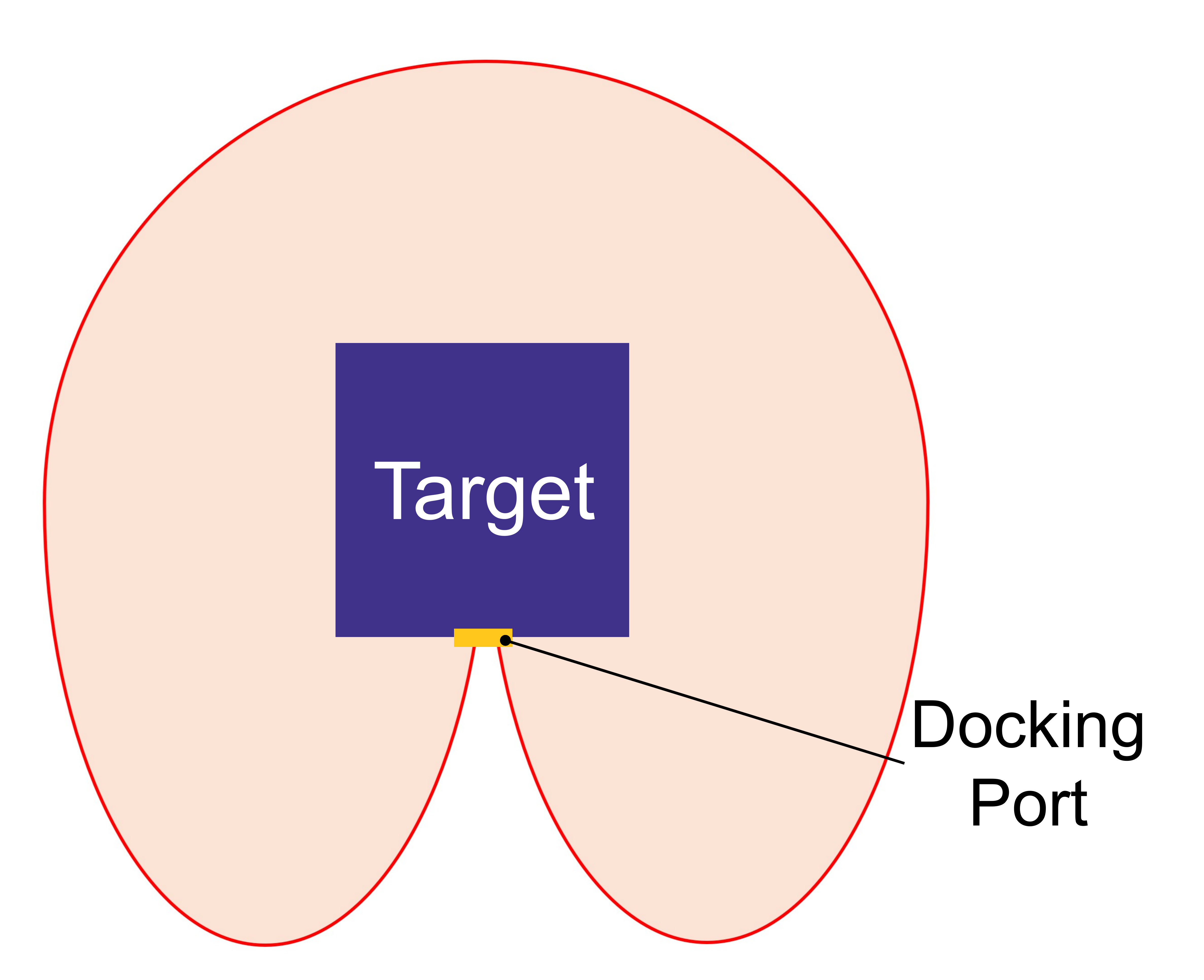}
  }
  \caption{Different KOS patterns changed depending on the relative pose between the target and the chaser.}
  \label{fig:KOS}
\end{figure}

As shown in \fig{fig:KOS}, the KOS dynamically changes depending on the state of the chaser \( \bm{x}_k \) and the state of the target \( \bm{p}_{\text{target},k} \).  
The configuration of the KOS is primarily determined by whether the chaser is located in front of or behind the docking surface of the target.  
In the rear region, a simple circular exclusion zone is imposed.  
In the front region, the KOS switches depending on whether the conditions for final approach are satisfied.  
Specifically, these conditions are met when both the angular deviation from the docking surface’s normal and the inter-center distance fall within specified thresholds.  
For the Keep-Out Sphere (KOS) design during the final approach phase, we refer to the KOS geometry proposed in the study by Richard Zappulla II et al.~\cite{zappulla2019}.
The distance threshold is set to $1.5\,r_{\text{safe}}$, and the angular threshold corresponds to the maximum allowable value that avoids collision between the corners of the satellites.

\subsubsection{State~I (General Approach Condition)}
The chaser is neither close to nor properly aligned with the target.  
In this state, the KOS constraints consist of a circular region of radius \( r_{\text{safe}} \) for basic safety and two half-elliptical regions (major axis: \( r_{\text{safe}} \), minor axis: \( r_{\text{safe}} / 2 \)) for final approach are imposed as KOS constraints.  
The definition of \( r_{\text{safe}} \) is provided in \eqref{eq:r_safe}.

\subsubsection{State~II (Final Approach Feasible Condition)}
The chaser is close to and properly aligned with the target.  
In this state, the circular constraint on the docking-surface side is relaxed, leaving only the elliptical KOS in effect, thereby allowing the final approach to proceed.

\begin{equation}
    r_{\text{safe}} = \frac{\sqrt{2}}{2} l_s + \frac{\sqrt{2}}{2} l_t + l_\text{margin}
    \label{eq:r_safe}
\end{equation}

As shown in \fig{fig:thruster_model}, $l_s$ denotes the side length of the chaser, and $l_t$ represents the side length of the target.  
The margin $l_{\text{margin}}$ is set to $10\%$ of $l_s$.  
This expression assumes the worst-case configuration in which the corners of the chaser and the target are closest, and defines the safety distance based on the minimum diagonal separation between the two cubes.

In State~II, the chaser is aligned with the target’s docking surface, and the most severe contact mode—corner-to-corner collision—is avoided.  
Accordingly, the KOS is adjusted to reflect this reduced risk.

\section{THRUSTERS CONTROL}
The optimal trajectory obtained in the previous section is based on continuous forces and torques (wrench) calculated at each time step $\Delta t$. However, most thrusters used in actual space missions are discrete actuators that can only be operated in an ON/OFF manner~\cite{esho2024, tummala2017}.
Therefore, in this section, we design a control system that enables the chaser to track the optimized trajectory using such ON/OFF thrusters.

\subsection{PD Control Law}
In this study, a proportional--derivative (PD) control law is adopted. 
At each time step $k$, we define the tracking error vector between the reference trajectory $\bm{x}_{\text{ref}}(k)$ and the simulated chaser state $\bm{x}_{\text{sim}}(k)$ as follows.

\begin{equation}
    \bm{e}(k) = 
    \bm{x}_{\text{ref}}(k) - \bm{x}_{\text{sim}}(k)
    =
    \begin{bmatrix}
    e_x & e_y & e_\theta & e_{v_x} & e_{v_y} & e_\omega
    \end{bmatrix}_k^\top
    \label{eq:pd_error}
\end{equation}

Using this error vector, the required wrench $\mathcal{F}_{\text{ref}}$ in the inertial (world) coordinate frame is computed as follows.

\begin{equation}
    \mathcal{F}_{\text{ref}}(k) =
    \begin{bmatrix}
    F_{x,\text{ref}} \\
    F_{y,\text{ref}} \\
    \tau_{\text{ref}}
    \end{bmatrix}_k
    =
    \begin{bmatrix}
    K_{p,P} \cdot e_x(k) + K_{d,P} \cdot e_{v_x}(k) \\
    K_{p,P} \cdot e_y(k) + K_{d,P} \cdot e_{v_y}(k) \\
    K_{p,A} \cdot e_\theta(k) + K_{d,A} \cdot e_\omega(k)
    \end{bmatrix}
    \label{eq:pd_control}
\end{equation}

Here,  
$\bm{K}_{p,P} $ and $\bm{K}_{d,P}$ are the proportional and derivative gains for position control, 
$\bm{K}_{p,A} $ and $\bm{K}_{d,A}$ are those for rotational (attitude) motion.

Next, the desired wrench $\bm{\mathcal{F}}_{\text{ref}}$ is transformed from the inertial frame to the body-fixed frame to enable thruster allocation. The translational force vector is rotated by the transpose of the rotation matrix $R(\theta_{\text{sim}})^\top$ as shown in ~\eqref{eq:rotation}. 

\begin{equation}
  \bm{\mathcal{F}}_{\text{b,ref}} = \bm{R}(\theta_k)^{\top} \bm{\mathcal{F}}_{\text{ref}}
  \label{eq:rotation}
\end{equation}

\begin{figure*}[t]
    \centering
    \begin{minipage}[t]{0.328\linewidth}
        \includegraphics[width=\linewidth]{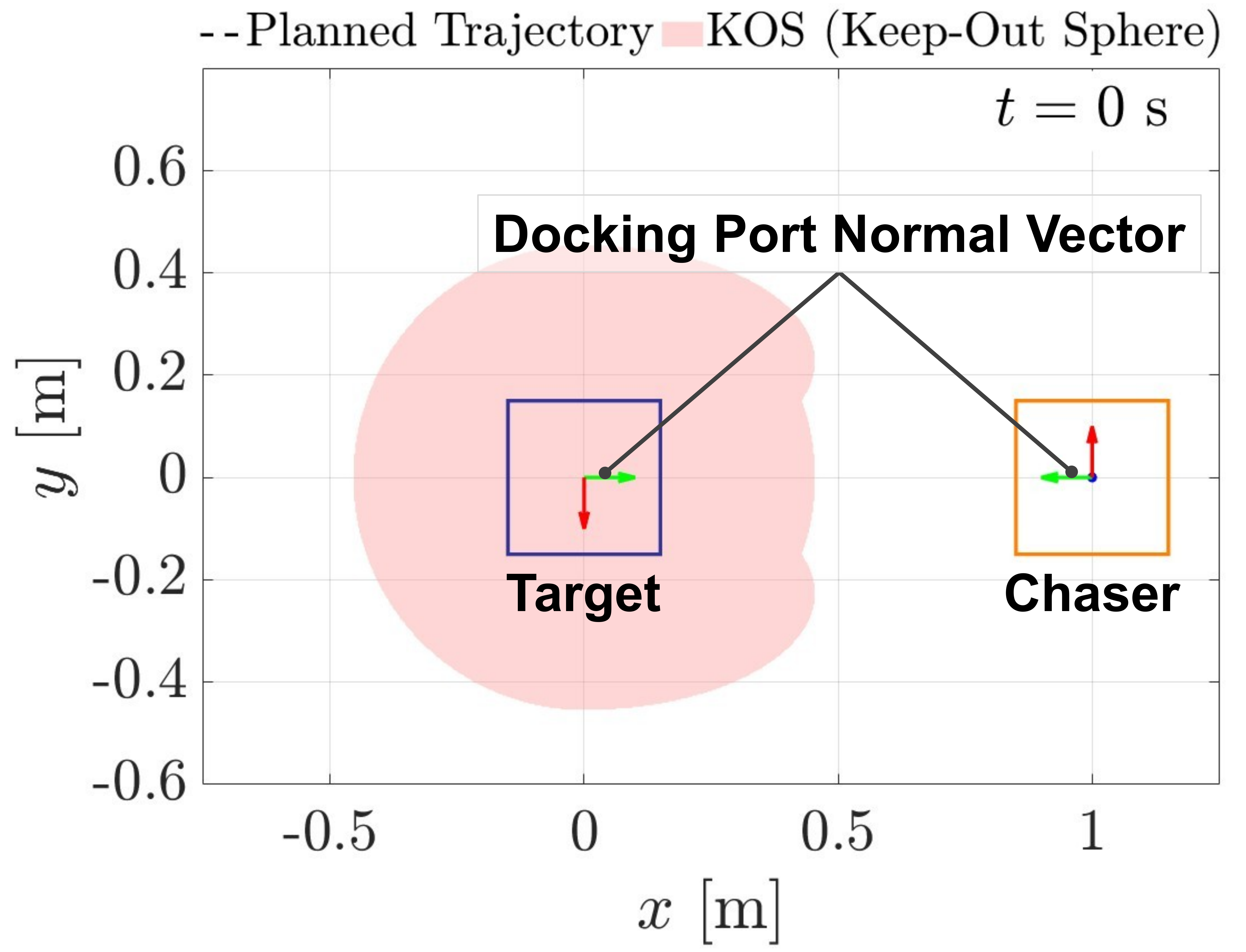}
    \end{minipage}
    \hfill
    \begin{minipage}[t]{0.328\linewidth}
        \includegraphics[width=\linewidth]{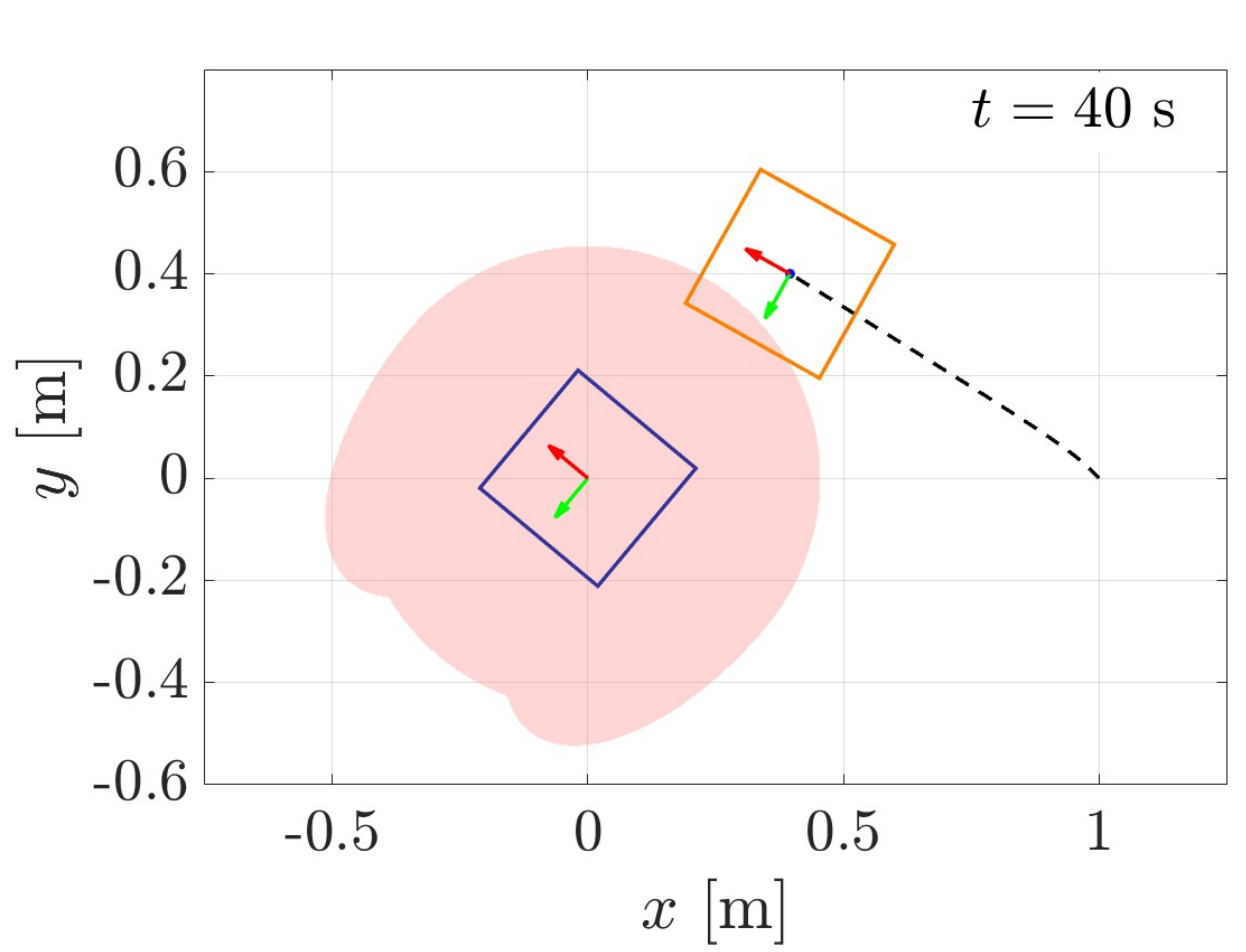}
    \end{minipage}
    \hfill
    \begin{minipage}[t]{0.328\linewidth}
        \includegraphics[width=\linewidth]{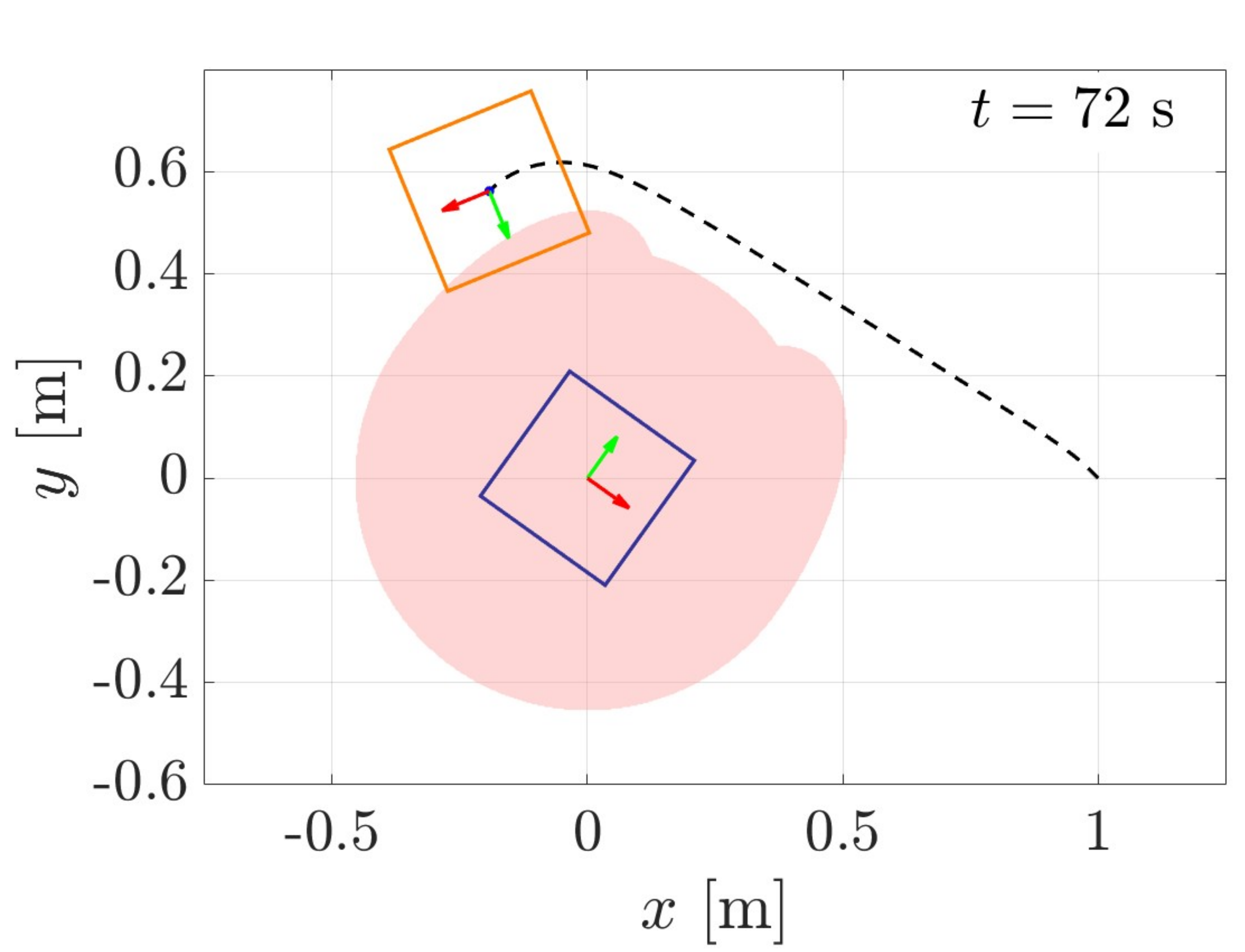}
        \vspace{-3mm}
    \end{minipage}
    
    \begin{minipage}[t]{0.328\linewidth}
        \includegraphics[width=\linewidth]{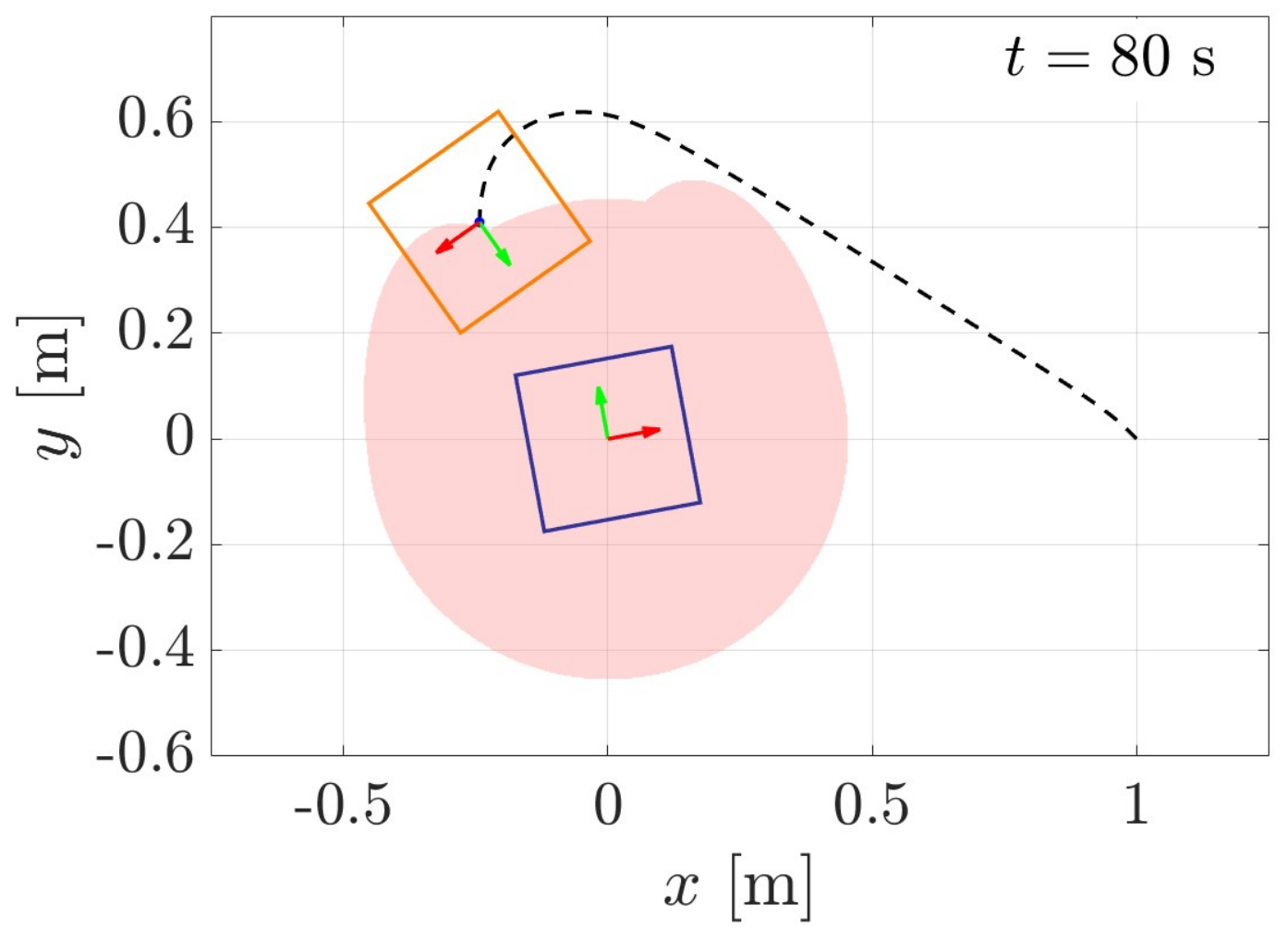}
    \end{minipage}
    \hfill
    \begin{minipage}[t]{0.328\linewidth}
        \includegraphics[width=\linewidth]{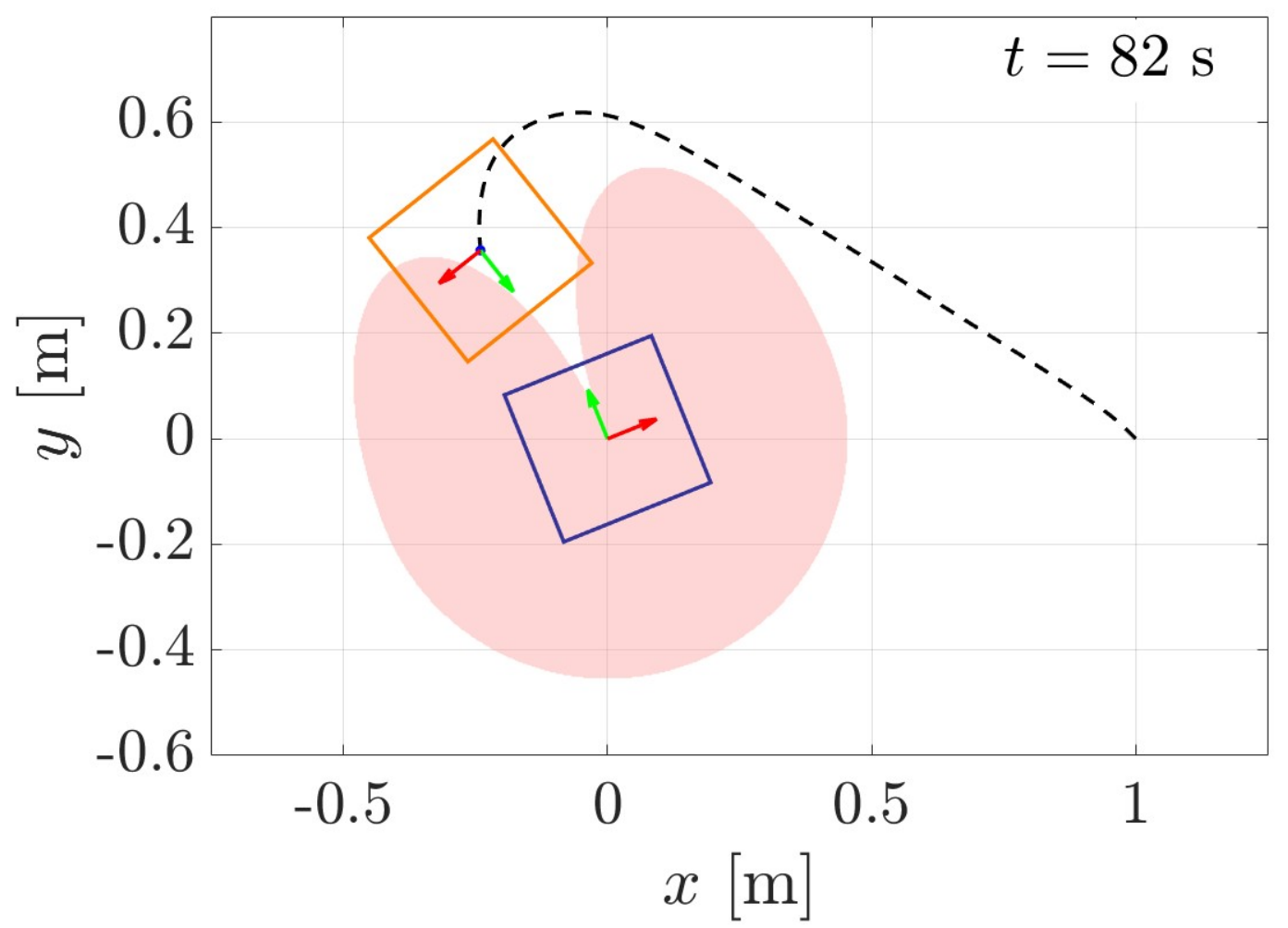}
    \end{minipage}
    \hfill
    \begin{minipage}[t]{0.328\linewidth}
        \includegraphics[width=\linewidth]{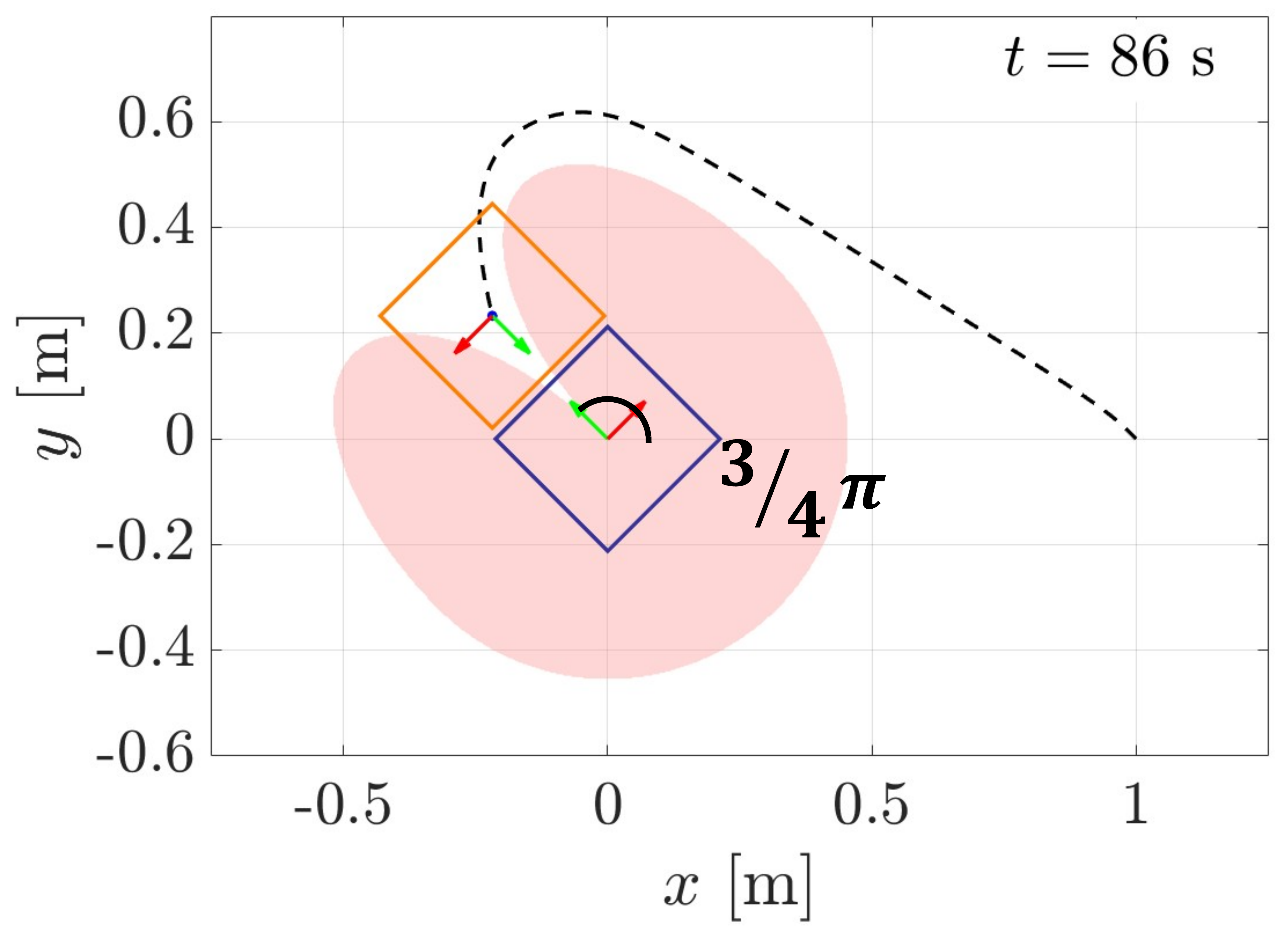}
    \end{minipage}

    \caption{Optimized trajectory for a closest approach (dot line) and the corresponding two satellites' poses in time series (two colored squares). In this case, the trajectory was planned to achieve the closest approach when the target is tilted at $3\pi/4$ to the $x$-axis.}
    \label{fig:Result_Opt}
\end{figure*}

\subsection{Thruster Output Allocation Strategy Using PWM Control}
To realize the desired wrench $\bm{\mathcal{F}}_{\text{ref}}$, the duty ratios $\bm{u} \in [0,1]^8$ of the eight ON/OFF thrusters mounted on the chaser satellite (as shown in \fig{fig:thruster_model}) are computed by solving the following linear least-squares problem:

\begin{equation}
  \min_{\bm{u}} \left\| \bm{B} \bm{u} F_{\text{thr}} - \bm{\mathcal{F}}_{\text{b,ref}} \right\|^2 \quad \text{s.t.} \quad 0 \le u_i \le 1
  \label{eq:thruster_allocation}
\end{equation}

Here, $\bm{B} \in \mathbb{R}^{3 \times 8}$ is the control effectiveness matrix that encodes the position and orientation of each thruster.

Once the optimal duty ratio $\bm{u}^\ast$ is obtained, the control interval $\Delta t$ is divided into $N_{\text{PWM}}$ sub-intervals, and pulse Width modulation (PWM) is applied to generate binary ON/OFF signals for each thruster accordingly.

\section{Numerical Experiments}
\subsection{Simulation setup}
In this section, to evaluate the effectiveness of the proposed method, we construct a simulation environment representing a two-dimensional microgravity space and perform trajectory optimization as well as PWM thruster-based tracking experiments under the following conditions:

\begin{itemize}
  \item The target satellite performs only rotational motion.
  \item The initial position of the chaser is set to $(x_0, y_0) = (1.0, 0.0)\,\mathrm{m}$.
  \item The target is modeled as a cube with a side length of $0.3\,\mathrm{m}$.
  \item The chaser is equipped with eight ON/OFF-type thrusters, each providing a constant maximum thrust defined per test condition.
  \item The simulation is implemented using MuJoCo~\cite{MuJoCo}, employing Euler integration with a time step of $\Delta t = 0.01\,\mathrm{s}$ and a control frequency of $10\,\mathrm{Hz}$.
  \item The weighting coefficients in~\eqref{eq:objective} were set to $w_{\text{goal}} = 100$ and $w_{u} = 10$.
\end{itemize}

The control cycle of the thrusters was determined based on considerations from free-floating experiments~\cite{santaguida2023} and actual satellite operations.

The weighting parameters were primarily determined empirically to ensure stable convergence of the optimization process.
The proposed optimization framework was also confirmed to converge under other sets of weighting parameters, and qualitative differences in the resulting trajectories were observed depending on the weight combinations.
However, a detailed sensitivity analysis on the influence of these parameters on the trajectory characteristics is beyond the scope of this paper.
The ratio between $w_{\text{goal}}$ and $w_{u}$ was chosen such that the magnitudes of the corresponding objective function terms are of the same order, thereby balancing the terminal position accuracy and the control effort.
These values were fixed throughout all case studies to maintain consistency in comparing the performance under different trajectory-generation conditions.

\subsection{Trajectory Planning}
We first verified that the proposed trajectory optimization method (Eqs.~\eqref{eq:objective}--\eqref{eq:collision_avoid}) successfully generates an ideal trajectory in the continuous control domain, allowing the chaser to approach the target while aligning its docking face. 
In this case, the approach attitude angle $\theta_{\text{approach}}$ of the target is set to $3\pi/4$, and its angular velocity is set to 0.1 rad/s. 
This angular velocity was intentionally chosen to be faster and more challenging than the reported typical rotation rate of actual space debris (e.g., rocket upper stages), which is approximately $0.06\,\mathrm{rad/s}$~\cite{vananti2023}.
As shown in the \fig{fig:Result_Opt}, the chaser safely approaches the target with high accuracy, maintaining proper face alignment while avoiding any violation of the restricted area. 
Furthermore, during the interval from $t = 80\,\mathrm{s}$ to $t = 82\,\mathrm{s}$, the constraints on relative distance and orientation between the chaser and the target are satisfied, allowing the restricted area to be updated so that the chaser can move closer to the target.

\subsection{Trajectory Tracking with the Thruster Control}
\begin{figure}[t]
    \centering
    \includegraphics[width=\linewidth]{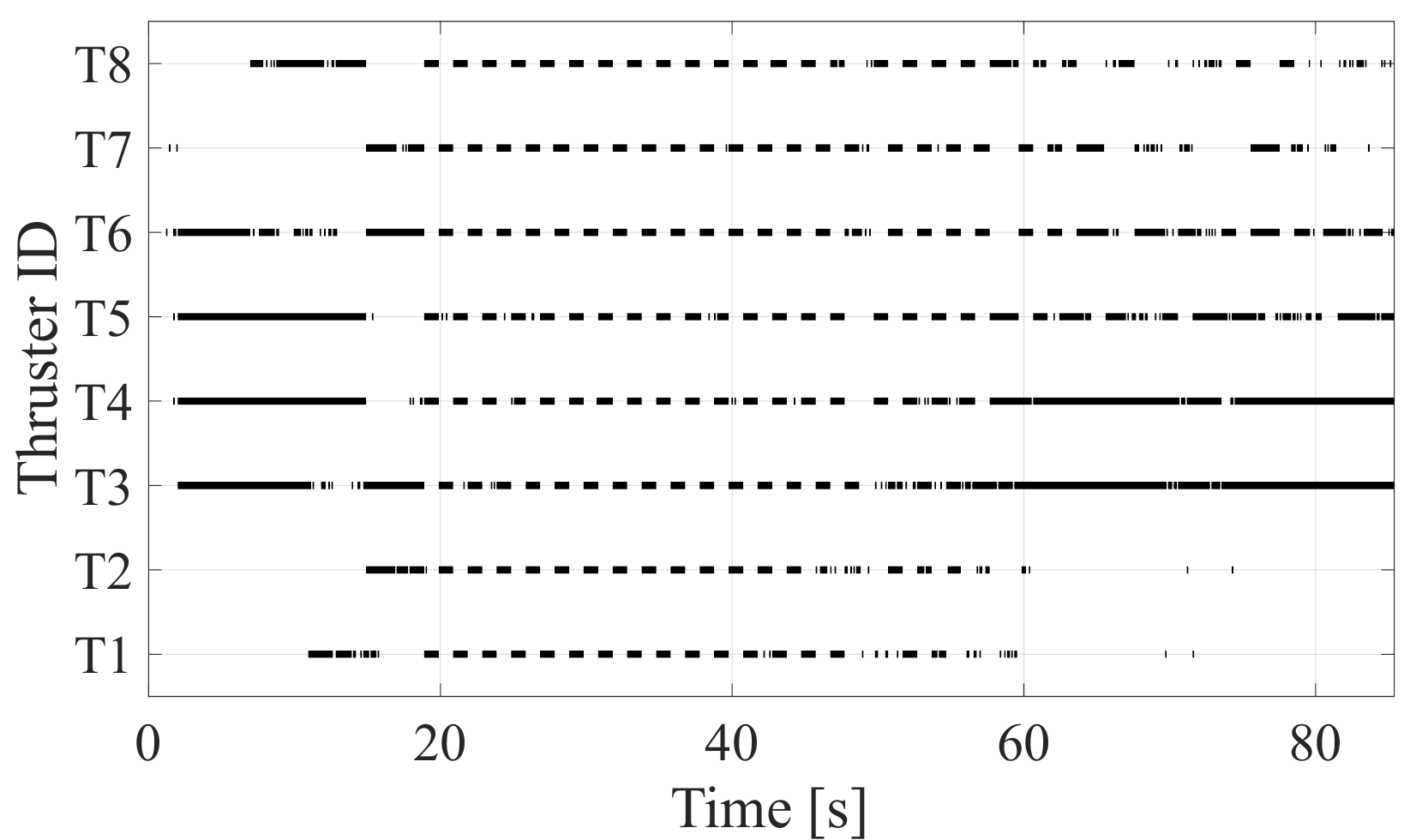}
    \caption{Thruster operation sequence (Control cycle: 10~Hz).}
    \label{fig:thruster_sequence}
\end{figure}

\begin{figure}[t]
    \centering
    \includegraphics[width=\linewidth]{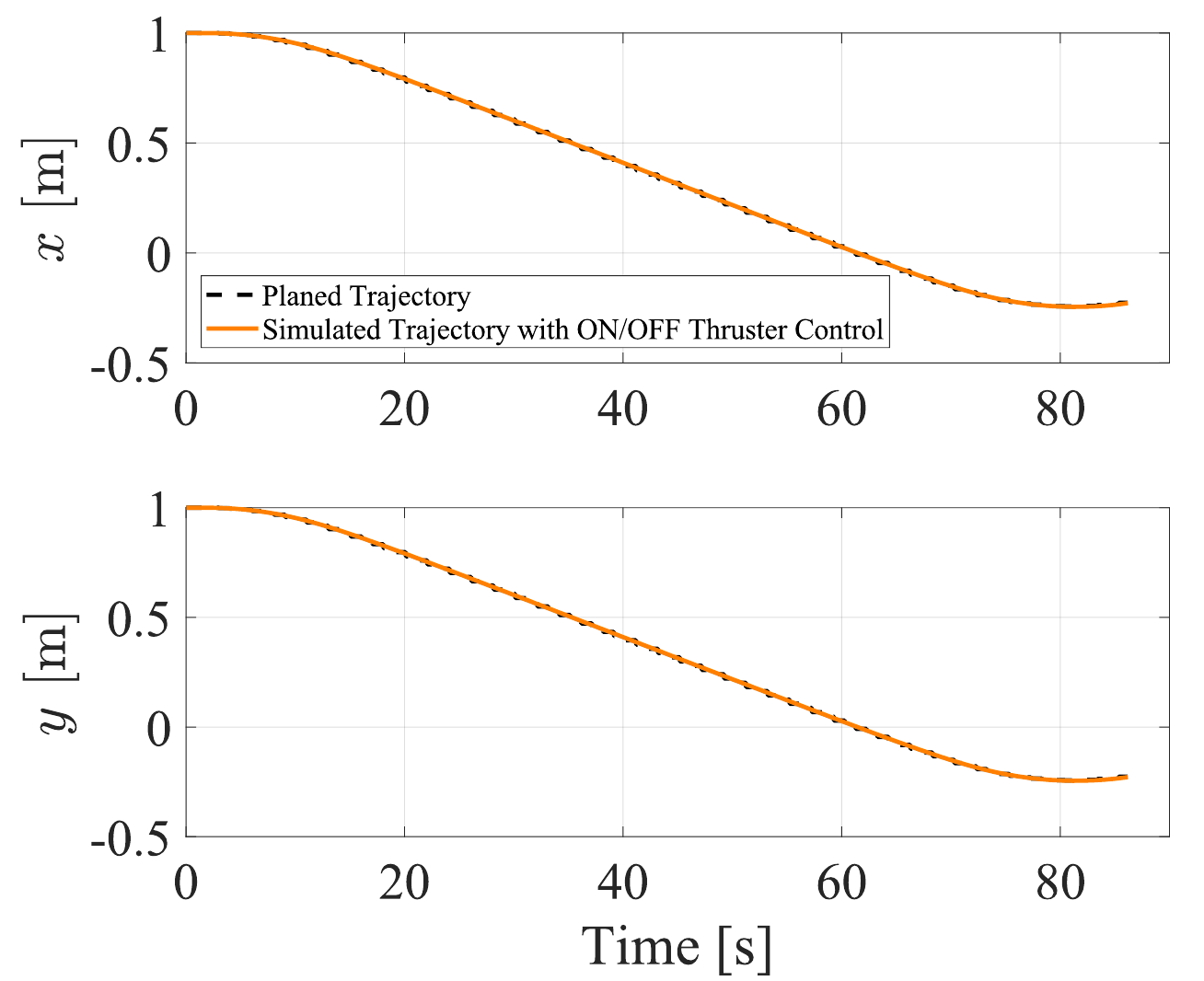}
    \caption{Trajectory tracking performance by the discrete thruster control of the chaser in MuJoCo simulation.}
    \label{fig:MuJoCo_Trajectory}
\end{figure}

We then verify the practical feasibility of the proposed method through numerical experiments in which the chaser satellite tracks the ideal trajectory using PWM-based thruster control. 
These simulations are conducted in a two-dimensional microgravity environment modeled in MuJoCo, and all control logic and simulation execution are implemented in Python. 
The discrete thruster system tracked the ideal trajectory generated under continuous control assumptions. \fig{fig:thruster_sequence} illustrates the resulting ON/OFF command sequences for the chaser's eight thrusters. 
The trajectory tracking performance is evaluated in a MuJoCo simulation (\fig{fig:MuJoCo_1}). 
Additionally, in \fig{fig:MuJoCo_Trajectory}, the solid line (actual trajectory) is compared to the dotted line (ideal trajectory), demonstrating that the chaser follows the optimized path with high fidelity. 

\begin{figure}[t]
    \centering
    \includegraphics[width=\linewidth]{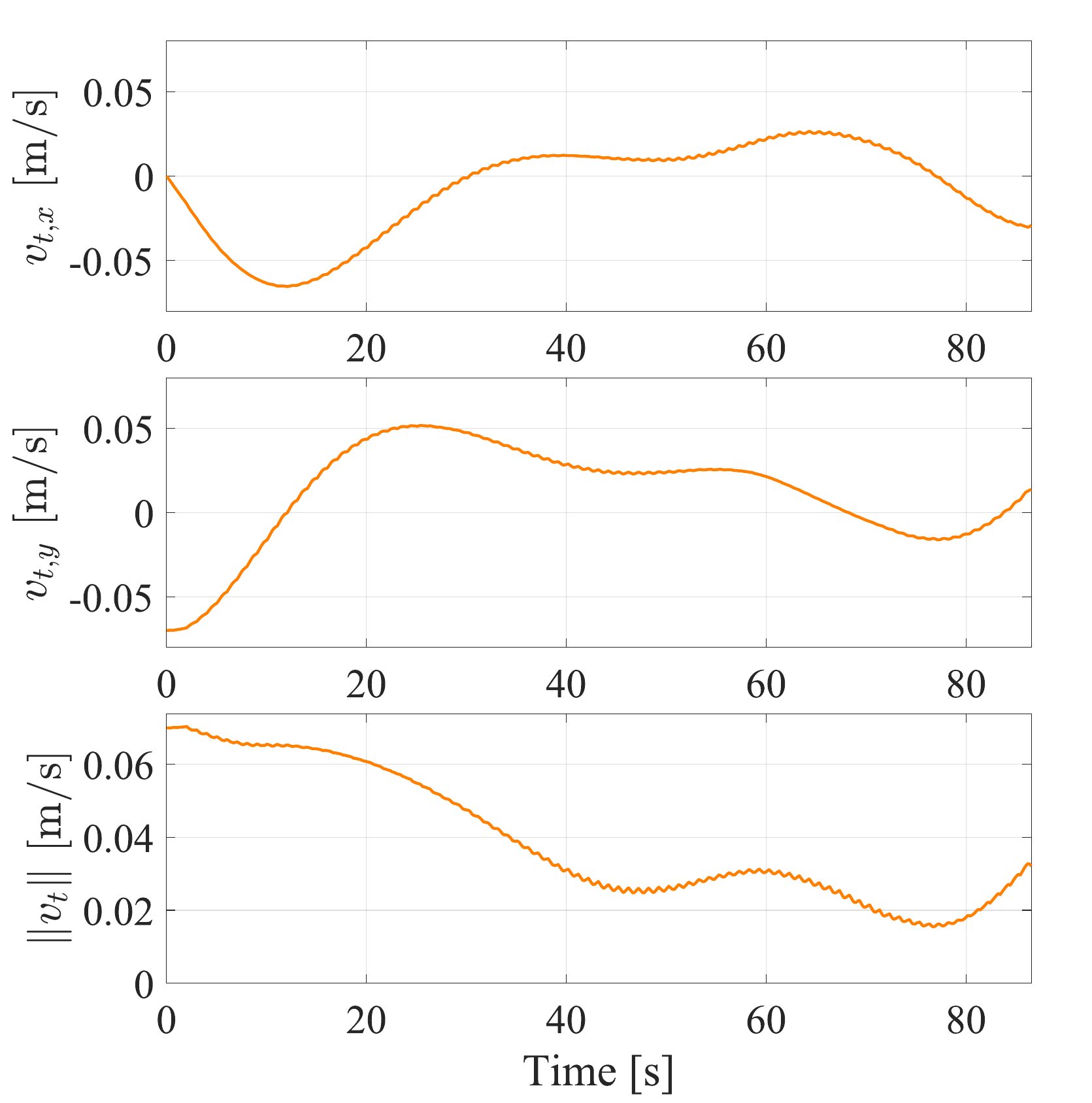}
    \caption{Time history of the chaser--target relative velocity in the target-fixed frame in MuJoCo simulation.}
    \label{fig:relative_velocity}
\end{figure}

To evaluate whether the chaser is synchronized with the rotating target also in terms of relative velocity, the time history of the chaser–target relative velocity in the target-fixed frame is shown in Fig.~\ref{fig:relative_velocity}.
The magnitude of the relative velocity converges to approximately $0.03\,\mathrm{m/s}$ at the end of the maneuver.
Although the optimization problem does not explicitly include the terminal relative velocity as an objective term, the combination of minimizing the chaser’s kinetic energy and terminal position error, together with attitude synchronization constraints, naturally results in a residual velocity of this order.
This terminal synchronization level is consistent with the capture feasibility thresholds reported in previous studies: Sasaki et al.~\cite{sasaki2023} defined $0.03\,\mathrm{m/s}$ as the allowable relative velocity for secure contact in an active debris removal mission, and Zhang et al.~\cite{zhang2024} achieved successful visual-servo-based capture at approximately $0.02\,\mathrm{m/s}$.
Therefore, the relative velocity obtained in this study demonstrates a comparable degree of dynamic synchronization with practical capture conditions.
While introducing an explicit terminal relative-velocity term could further reduce the residual motion, the current formulation provides a well-balanced trade-off between terminal accuracy and relative-motion stability, and its simplicity ensures consistent applicability across the subsequent case studies.

\subsection{Case Study}
Next, we present the results of trajectory planning using the proposed optimization method under various conditions. 
The evaluation settings and corresponding results for each case are summarized below.

\vspace{2 mm}
\subsubsection{Case Study 1): Variation in Target Spinning Velocity}
In this case study, we evaluate the performance of trajectory planning by varying the angular velocity of the target satellite ($\omega_{\mathrm{target}}$) from $0.035$ to $2.0\,\mathrm{rad/s}$ in increments of $0.025\,\mathrm{rad/s}$.  
For each angular velocity, the thruster output force $F_{\mathrm{thr}}$ is varied from $0.03$ to $1.02\,\mathrm{N}$ in increments of $0.03\,\mathrm{N}$.

\begin{figure}[t]
    \centering
    \includegraphics[width=\linewidth]{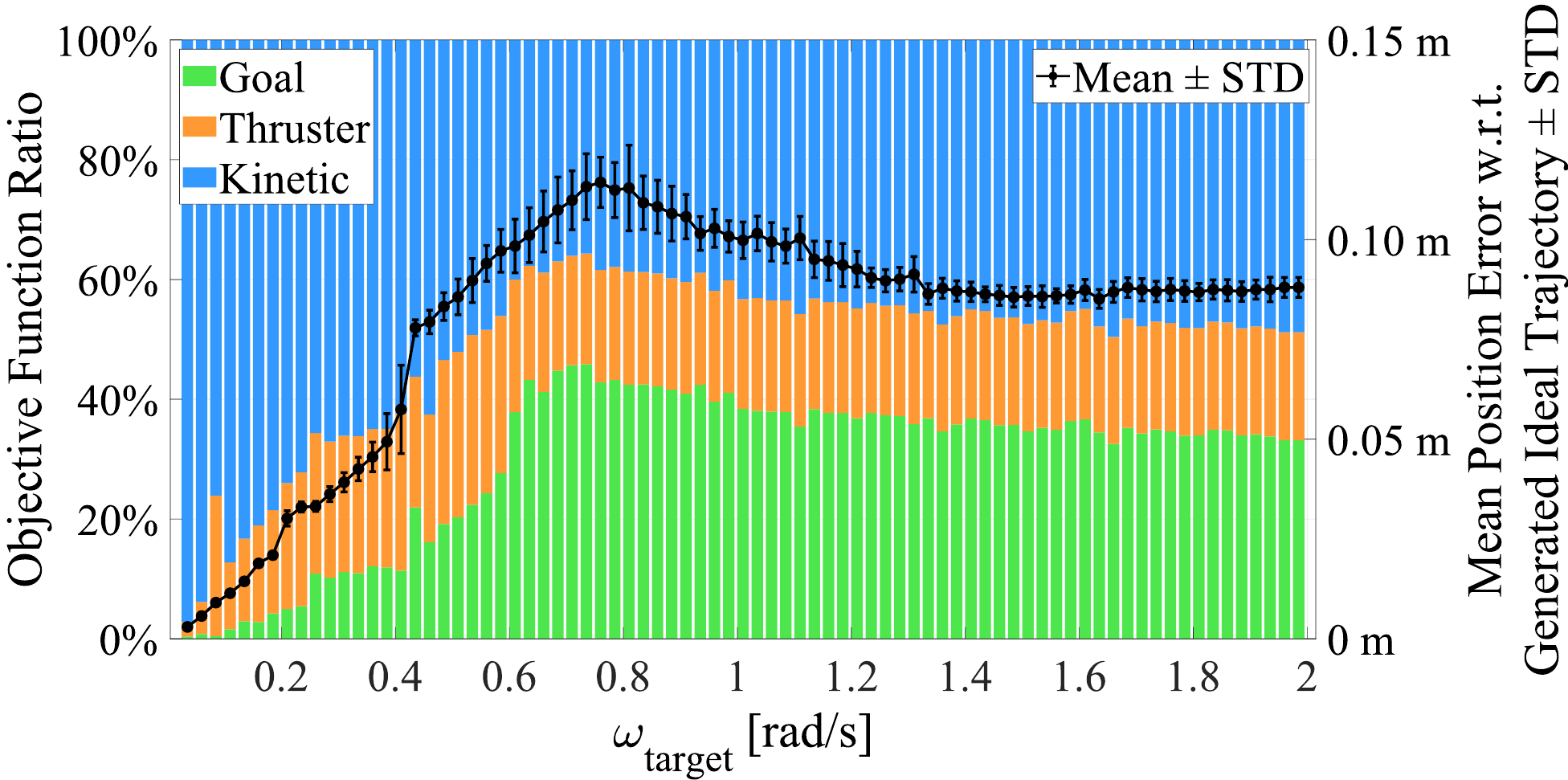}
    \caption{Effect of target angular velocity ($\omega_{\mathrm{target}}$) on final position error. }
    \label{fig:case1_result}
\end{figure}

The results are presented in \fig{fig:case1_result}. 
This figure employs a dual-axis overlay format, where the horizontal axis corresponds to the target's angular velocity $\omega_{\mathrm{target}}$. 
The left vertical axis depicts the relative contributions of the three components of the objective function—namely, the goal tracking term, the control effort term, and the kinetic energy term—visualized as a stacked bar chart. 
Superimposed on this, the right vertical axis shows the mean final position error and its standard deviation (Mean ± STD), averaged across all thruster output conditions, using an error bar plot.

The final position error reaches its peak when the target's angular velocity, $\omega_{\mathrm{target}}$, is approximately $0.75\,\mathrm{rad/s}$, then gradually decreases until around $1.25\,\mathrm{rad/s}$, and eventually converges to a steady-state error of approximately $0.09\,\mathrm{m}$.  
This behavior can be attributed to a change in the trajectory shape: when the angular velocity exceeds $0.75\,\mathrm{rad/s}$, the chaser modifies its approach path to rotate around the target in the same direction as the target's rotation before the final approach.  
This change in trajectory leads to an increase in the kinetic energy term in the objective function.  
In other words, the penalty associated with the final position error becomes more significant than the penalty for increased kinetic energy.  
As a result, the optimal control strategy prioritizes suppressing the final position error by applying larger control forces, even at the expense of higher energy consumption.  
In particular, when the target's angular velocity reaches approximately $0.75\,\mathrm{rad/s}$, the dominant term in the cost function transitions from the kinetic energy term to the final position error term.  
This shift is also evident in the bar graph shown in \fig{fig:case1_result}.


\begin{figure}[t]
    \centering
    \includegraphics[width=.9\linewidth]{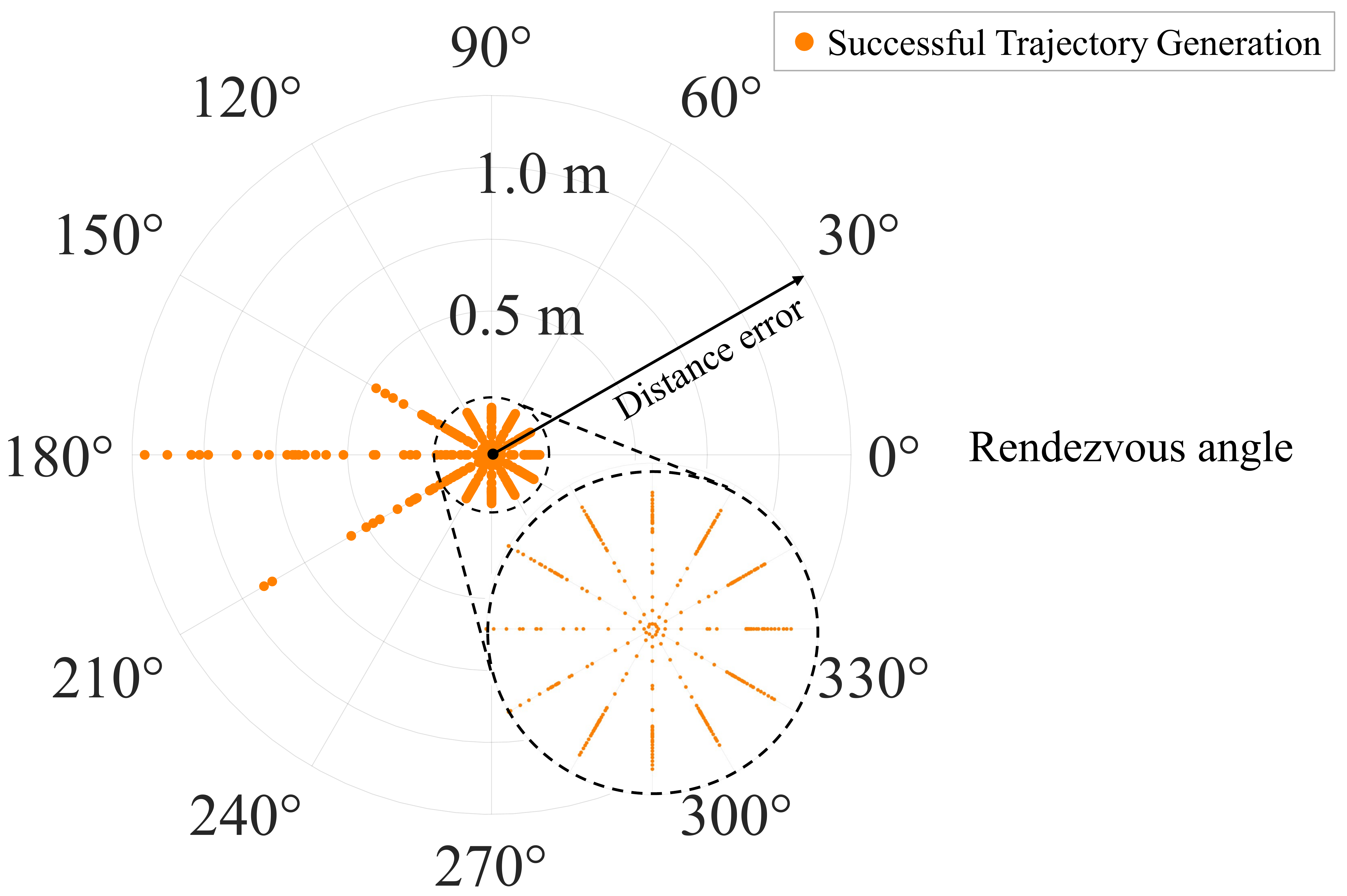}
    \caption{Effect of target attitude angle ($\theta_{\mathrm{approach}}$) on final position error. (Final attitude error consistently converges to a value near zero.)}
    \label{fig:case2_result}
\end{figure}

\vspace{2 mm}
\subsubsection{Case Study 2): Variation in Target Attitude at Closest Approach}
In this case study, we vary the target’s attitude angle $\theta_{\mathrm{approach}}$ at the point of closest approach from $0^\circ$ to $330^\circ$ in $30^\circ$ increments.  
For each attitude angle, the target's angular velocity $\omega_{\mathrm{target}}$ is varied from $0.05$ to $2.00\,\mathrm{rad/s}$ in increments of $0.05\,\mathrm{rad/s}$, while the thruster output force, $F_{\mathrm{thr}}$, is fixed at $0.03\,\mathrm{N}$.

The results are presented in \fig{fig:case2_result}. 
In this figure, the target's attitude angle $\theta_{\mathrm{approach}}$ is represented as the angular coordinate in a polar plot, while the radial coordinate corresponds to the final position error. 
Hence, a greater distance from the origin indicates a larger final error.

It was observed that the final position error is significantly higher for conditions where $\theta_{\mathrm{approach}} = 150^\circ$ to $210^\circ$, compared to other configurations.
This indicates that when the chaser attempts to approach from the rear side of the target, the final position error tends to increase. 
In particular, at $\theta_{\mathrm{approach}} = 180^\circ$, which corresponds to a direct approach from the rear, a pronounced increase in error is observed.

In both Case Study 1) and 2), the final attitude is imposed as a hard constraint in the optimization.  
As a result, whenever an ideal trajectory is successfully generated, the final attitude error consistently converges to a value near zero (at most on the order of $10^{-33}$). 
The evaluation in these case studies was limited to the trajectory generation stage, as the trajectory in thruster-based control is primarily achieved via PD control, whose performance heavily depends on the choice of gains, making a pure performance assessment difficult.

\section{Conclusion}
\label{sec:conclusion}
In this study, we proposed a trajectory optimization and ON/OFF thruster-based tracking method for a chaser spacecraft to approach and align with a target in a two-dimensional, microgravity environment. 
The proposed method consists of two stages: first, an ideal trajectory is generated via nonlinear optimization under the assumption of continuous control inputs; second, this trajectory is tracked using pulse-width-modulated (PWM) ON/OFF thrusters.

We first verified that the proposed optimization method successfully generates trajectories in which the chaser approaches the docking face of the target safely and accurately, while avoiding restricted zones and satisfying the chaser’s thrust constraints.  
Furthermore, by introducing dynamically updated restricted regions based on the relative distance and orientation between the chaser and the target, we demonstrated that the chaser can approach to very close proximity during final alignment while reducing the risk of collision.

Next, through two-dimensional simulations using MuJoCo, we confirmed that the trajectories can be tracked with high precision using a more realistic ON/OFF thruster model.

In addition, we evaluated the characteristics and robustness of the proposed method through two case studies.  
In Case Study 1), we systematically varied the target's angular velocity and the chaser's thruster output, and evaluated the resulting trajectories. 
The results showed that the trajectory generation remained feasible with acceptable error levels for target angular velocities up to approximately $2.0\,\mathrm{rad/s}$.  
In Case Study 2), we analyzed how the final position error varied with the target’s approach attitude $\theta_{\mathrm{approach}}$. 
It was found that the error significantly increased when the chaser approached from the backside of the target, particularly in the range $\theta_{\mathrm{approach}} = 150^\circ$ to $210^\circ$, with the maximum error observed around $180^\circ$, corresponding to the target's rear side.

These results demonstrate that the proposed trajectory planning and control framework can maintain high accuracy and safety under diverse mission conditions. 
Moreover, the analysis identified specific approach scenarios in which tracking becomes significantly more difficult. 
Future work will focus on application in the three-dimensional tumbling case and the real-world hardware experiments using a two-dimensional air-floating testing platform.


\bibliography{./IEEEabrv,bibliography.bib}

\end{document}